# Deep Labeling of fMRI Brain Networks Using Cloud Based Processing

Sejal Ghate[1], Alberto Santamaria-Pang,[2] Ivan Tarapov, [2] Haris I Sair, [3,4] and Craig K Jones[3,4,5]

[1] Department of Biomedical Engineering, Johns Hopkins University, Baltimore MD 21218, USA

[2] Health and Life Sciences, Microsoft, Redmond Washington

[3] Department of Radiology and Radiological Sciences, Johns Hopkins School of Medicine, Baltimore MD 08544, USA

[5] Malone Center for Engineering in Healthcare, Johns Hopkins University, Baltimore MD 21218, USA

[5] Department of Computer Science, Johns Hopkins University, Baltimore MD 21218, USA

**Abstract**

Resting state fMRI is an imaging modality which reveals brain activity localization through signal changes, in what is known as Resting State Networks (RSNs). This technique is gaining popularity in neurosurgical pre-planning to visualize the functional regions and assess regional activity. Labeling of rs-fMRI networks require subject-matter expertise and is time consuming, creating a need for an automated classification algorithm. While the impact of AI in medical diagnosis has shown great progress; deploying and maintaining these in a clinical setting is an unmet need. We propose an end-to-end reproducible pipeline which incorporates image processing of rs-fMRI in a cloud-based workflow while using deep learning to automate the classification of RSNs. We have architected a reproducible Azure Machine Learning cloud-based medical imaging concept pipeline for fMRI analysis integrating the popular FMRIB Software Library (FSL) toolkit. To demonstrate a clinical application using a large dataset, we compare three neural network architectures for classification of deeper RSNs derived from processed rs-fMRI. The three algorithms are: an MLP, a 2D projection-based CNN, and a fully 3D CNN classification networks. Each of the networks was trained on the rs-fMRI back-projected independent components giving >98% accuracy for each classification method.

## 1    Introduction

Functional magnetic resonance imaging (fMRI) is a technique to understand time-varying, spatially related signal changes in the brain. Since its introduction, fMRI has revolutionized our understanding of neuroscience and human cognition [1]. fMRI relies on Blood Oxygen Level Dependent (BOLD) signal change that is modulated by oxygen uptake in functionally active regions of the brain [2]. Resting state fMRI (rs-fMRI) is a method to measure intrinsic signal fluctuations without a specific task-based paradigm and is gaining popularity for brain activity localization in neurosurgical pre-planning such as for the eloquent cortex in the brain [3]. The benefits of using rs-fMRI are that highly correlated brain networks/components at rest can be reproduced across multiple subjects for appropriate assessment of functional brain regions. Moreover, rs-fMRI eliminates the need for active patient participation and other task-related logistical/demographic considerations such as cognitively challenged patients for language mapping.

Independent Component Analysis (ICA) [4] is a mathematical algorithm to separate 3D+time rs-fMRI data into 3D spatial maps of statistically independent components (ICs) having strong 1D temporal coherence. ICA is advantageous in that it is a blind-source separation technique that does not require a-priori information for clustering of temporal signals. Group-ICA is a popular method for performing ICA across a cohort to identify Resting State Networks (RSNs) common to the group as opposed to subject-level ICA which separates actual neuronal signals from noise for a single subject. Labeling of these spatial-temporal RSNs can be time consuming and subjective requiring expertise, thus creating the need for an objective and accurate algorithm for automated classification.

Over the past few years, deep learning has been gaining popularity to classify relevant RSNs from rs-fMRI ICA results. Kam et. Al. proposed a novel spatial-temporal deep-learning framework to identify noise components from true RSNs, using a 3D CNN for spatial ICA maps and 1D CNN on ICA time series [5]. Other studies such as Vergun et. Al., Zhao et al. used deep learning for RSN classification of a smaller number of ICs (~5-10) [6,7]. Deep learning based rs-fMRI classification was also used for disease classification in Alzheimer's disease and schizophrenia [8,9]. Our work is similar to Joliot et al's in investigating neural network for classifying ICA signals for a greater number of RSNs (>40) [10]. Though, in their work, subject specific contributions from group-ICA results were not used as part of their deep learning training data. In our study, we use the group-ICA results back-projected to individual subjects to highlight variability of the same RSN across different subjects, while also classifying a higher number of RSNs (58) displaying deep networks.

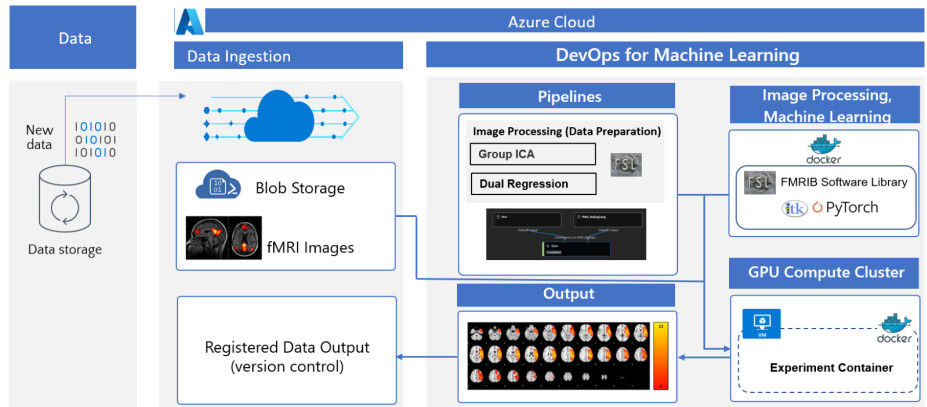

**Figure 1**: Schematic of the Concept Imaging Pipeline.

Our goal was to architect a reproducible Azure Machine Learning [11] cloud-based medical imaging concept pipeline for fMRI analysis integrating the popular toolkit FMRIB Software Library (FSL) [12] and to use it for a reproducible end-to-end image processing and deep-learning rs-fMRI classification framework. Figure 1 shows a schematic of our concept pipeline. This workflow incorporates both group-ICA processing and classification of RSNs on a single platform without the hassle of individual software processing and expert labeling. To the best of our knowledge, this is the first study that performed classification on over 50 RSNs and that will enable the inclusion of a hierarchy in the labeled networks. We plan to make of public access our pipeline architecture code and methods for further reproducibility.

## 2 Methods

### 2.1 Data

To demonstrate feasibility and reproducibility, we used the 1000 Functional Connectomes Project publicly available dataset for our experiments. The 176 subjects (106M/70F) from the Beijing-Zhang cohort were selected (mean age of 21.2±1.9) all data acquired on a 3T scanner, and the fMRI data had a TR = 2s, more details on the projects website [17].

### 2.2 fMRI Resting State Analysis

The rs-fMRI-analysis was composed of three main steps: (1) standard pre-processing pipeline to resample image volumes to a standard reference coordinate system; (2) performing group ICA to estimate group level components; and (3) dual regression analysis to estimate group components at the subject level.

Step 1: The pre-processing pipeline consisted of several steps implemented using FSL [13] including motion correction (MCFLIRT with standard parameters), spatial smoothing (FWHM=7mm), temporal filter (default high pass filter), and registration (resampling to MNI standard space using FLIRT).

Step 2: Group ICA of rs-fMRI was performed using FSL MELODIC. The MELODIC model order was set to 100 ICs and was selected based on highest variability amongst respective ICs. ICA output maps were reviewed and labeled as RSNs based on anatomic location by a neuroradiologist with 12 years of fMRI expertise. There were 58 unique labels identified for classification. Two classes, 'Noise' and 'Unknown' were used for ICs that did not display true RSNs.

Step 3: We used the dual regression (back-projection) algorithm implemented in FSL to estimate group components per subject. This was implemented in two stages. Stage 1, regress the group ICA spatial maps (from Step 2) into each subject's 4D dataset to resulting in a set of time courses; and Stage 2, regress the time courses into the same 4D dataset to get a subject-specific set of spatial maps [14]. All the processing components were executed using Azure Machine Learning Cloud-based analytics.

### 2.3 Neural Network Methods

The total dataset size from the back-projected dual regression processing was 17,600 3D volumes of size 45*54*45 voxels each. The data was split into training, validation, and testing groups (70/10/20%) and the split was performed at the subject level. Each split had a similar distribution of labels representative of typical rs-fMRI group-ICA results. Three neural network architectures were compared in terms of accuracy, time to train, and time to predict. To account for the class imbalance, where 'Noise' dominated as compared to the 57 other classifications, all labels were weighted according to their distribution in the dataset. The SGD optimizer, and Cross Entropy loss with weighted labels, were used for all three networks. Data was trained with a learning rate of 1e-3, a batch size of 32 and 25 epochs across all three networks. All training was performed on the AzureML platform using an NVIDIA K80 GPU compute cluster.

**MLP**: We chose to first incorporate a multi-layer perceptron (MLP), as opposed to CNNs, to understand the voxel-voxel interactions through the fully connected layers. Three fully connected hidden layers were used of 200 neurons each, with ReLU activation and a dropout of 66% after the second layer. No down sampling or data augmentation was performed on the dataset before training.

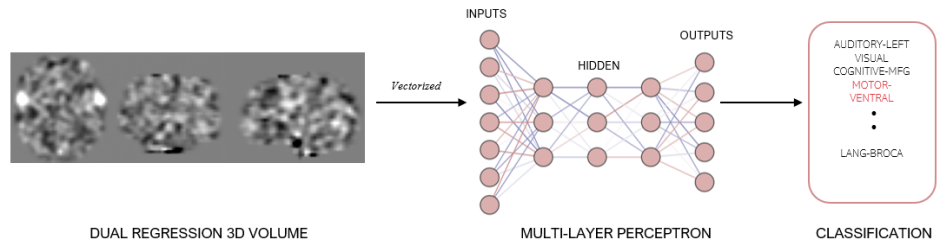

**Figure 2:** Pipeline of the classifications: (left) Axial, Coronal and Sagittal images of a 3D back-projected rs-fMRI dataset is flattened and input to the (middle) Multi-Layer Perceptron consisting of 3 hidden layers of 200 neurons each, and finally (right) classifying a specific RSN label.

**2.5D Neural Network:** The novel 2D data representation was constructed from the back-projected 3D volumes transformed into 3-channel 2D slices by projecting the sum of the voxel information across each of the axial, sagittal and coronal planes and setting them in the red, green and blue channel, respectively. Each of the projected image intensities were scaled to an RGB intensity range (0-255). This transformation thus retained all the information yet drastically reduced input volume size. A 2D ResNet (resnet-50 in PyTorch) with pretrained weights was trained on these images using the same hyper-parameters as the previous model.

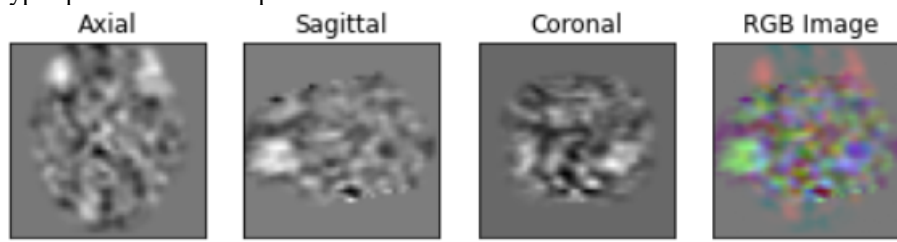

**Figure 3:** The input of the 2.5D network is a constructed RGB image based on (left) an axial projection of the back-projection of the fMRI IC (red channel), (middle-left) sagittal projection (green channel), (middle-right) coronal projection (blue), and (right) the 2.5D resulting RGB image.

**3D Neural Network**: A pretrained 3D ResNet architecture (r3d_18 in PyTorch) from the Torchvision video models was trained on the 3D fMRI cluster volumes. No additional data augmentation was performed on the datasets prior to training.

## 3    Results

### 3.1    fMRI Resting State Analysis

We architected a reproducible Azure Machine Learning [11] cloud-based medical imaging concept pipeline for fMRI analysis integrating the popular toolkit FMRIB Software Library (FSL). The core components of the concept pipeline are: 1) data ingestion component using a Blob Storage Container [15]; 2) Docker container with Ubuntu 18.04, FSL v6.0 [12] and Miniconda [16]; 3) a Python script to run FSL; and 4) configurable compute cluster. The pipeline is capable to orchestrate end-to-end steps or individual steps based on the configuration. We integrate modularity by building docker containers for required processing steps, one to run FSL-based image analytics and another Docker container built around the PyTorch image libraries. To enable full reproducibility, the pipeline automatically tracks datasets, code version control and Anaconda environments (and code dependencies). Similarly, the analysis pipeline has full traceability of input parameters for downstream analysis. All the algorithms packaged in docker containers and executed from the Azure Machine Learning Cloud. Figure 2 below shows an example of an image registration output using FLIRT (pre-processing step) when executed in the pipeline. The image in gray is the rs-fMRI volume (spatial component), red contours correspond to the reference image. Once individual dual regression volumes were estimated, they were used to train a classifier to semantically predict the resting state components.

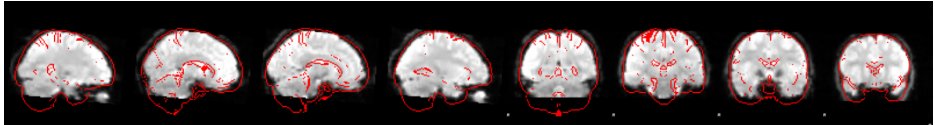

**Figure 4:** Example of a representative registration of a subject (grayscale) and the reference image (red contours).

### 3.2 Neural Network Performance Comparison

Our results indicate that all three neural networks could be used as an accurate classifier to classify the RSNs, showcasing greater than 98% testing accuracy in all models. Since the hyperparameters used were the same across all models, and no data augmentation was performed on the training dataset, a comparison was conducted based on accuracies obtained and time taken to train the model for optimization purposes. Table 1 below summarizes the training, testing accuracies and duration of model training and inference.

| Model | Training Accuracy | Testing Accuracy | Training Duration | Inference Duration |
|-------|-------------------|------------------|-------------------|--------------------|
| MLP   | 99.8%             | 100%             | 5 mins            | 1.9 s              |
| 2.5D  | 99.4%             | 99%              | 33 mins           | 23 s               |
| 3D    | 98.1%             | 98%              | 12 hrs 23mins     | 904 s              |

**Table 1:** Summary of the training and testing rs-fMRI classification accuracy and total training time taken.

As shown in Table 1 above, the MLP neural network reached a 100% testing classification accuracy, while the 2.5D and 3D models reached 99% and 98% testing classification accuracies.

Another important factor is the time to train each network considering future planned work with larger datasets. Training and inference were fastest for the MLP (5 minutes and ~2 seconds, respectively), while the longest training and inference time was taken by the 3D CNN model training (over 12 hours for training and 15 minutes for inference).

## 4 Discussion

The purpose of training our dataset on three different models was to understand which model would be most efficient for deployment and resulted in accurate classifications. Though all models gave excellent accuracies, the MLP had a 100% accuracy on the test data and the shortest inference time. Though CNN architectures are more commonly used for image classification problems, we believe that the MLP was able to extract voxel-to-voxel interaction at a lower level as opposed to high-level features such as edges and contours derived from a CNN. This reasoning can also be validated looking at the back projected rs-fMRI images which do not display any strong edges or contours when qualitatively and visually assessing the image. Additionally, since only 200 neurons were used in the hidden layers, the number of parameters would be considerably fewer than present in a 3D-CNN network. Hence, we infer that the ideal model to be used for further studies would be the MLP given the fast inference time for deployment purposes and similar accuracy to the convolutional architectures.

Another factor that may have led to such high accuracies is the distinct variability amongst different networks and the deep labeling performed in the dataset. Labels were annotated in such a way that each label comprised of a functional name (e.g., 'Language', 'Motor', or 'Visual') along with the anatomical region of the functional network (e.g., 'Ventral', 'Dorsal', 'Superior', etc.). Table 2 displays examples of the labels used in the classification. The purpose of this sequence adopted during the labeling is the emergence of certain hierarchies between sub-networks that can also be easily distinguished while classification. The labels displayed show lowest-level specific brain networks.

| DMN-PCC-MID | EXECUTIVE-POSTERIOR-LEFT |
|---|---|
| ATTENTION-DORSAL-IPS-MID | MOTOR-VENTRAL |
| VISUAL-LINGUAL-ANTERIOR | SENSORY-DORSAL-HAND-RIGHT |
| DMN-CINGULATE-MID | SALIENCE-INSULA-POSTERIOR |
| COGNITIVE-MFG | LANG-BROCA |

**Table 2:** Example labels of rs-fMRI networks.

Since the current study incorporates back-projected data from a group-ICA processing conducted for a model order of 100 independent components, the RSNs produced account for ICs that may display a deeper classification of RSNs compared to a group-ICA study with model order 20. As an extension of the current study, we aim to perform group-ICA on the cloud-based FSL pipeline for different predefined independent component model orders, to generate greater variability in the dataset which may include higher level, bigger RSNs as well as deeper, more specific RSNs. Additionally, the current dataset only includes one demographic of data acquired from Beijing. With the FSL image processing pipeline in the cloud, we plan to conduct an experiment on all subjects across every cohort present in the 1000 Functional Connectomes dataset, creating a dataset that is representative across different demographics.

The inclusion of RSNs with different levels of hierarchies and comprehensive labeling enables the extension of this study not just for multi-class classifications but also for multi-label classification providing information across taxonomies of functional networks and their respective anatomical regions. This would be particularly beneficial to identify two ambiguous RSNs that may overlap with each other as a result of coherence during ICA separation. The results from the deep learning classifications can be used to further create a hierarchical classification model.

In realistic scenarios, end-to-end cloud-based systems must be capable of not only training and deploying AI models but to pre-process large amounts of data using computational tools that were not originally designed to be operated in the cloud. This scenario poses the challenge of how to architect robust cloud systems so that they benefit from new capabilities from cloud computing while they can efficiently integrate standardized medical imaging libraries (such as FSL) to provide an end-to-end cloud-based pipeline. We plan to make our pipeline core components publicly available and further optimize the integration of FSL and enable parallel processing to minimize the processing time.

## 5      Conclusions

We have demonstrated interoperability of well-established image processing libraries with state-of-the art cloud-based architectures for a large dynamic imaging modality like fMRI. We have efficiently processed rs-fMRI image data by integrating image processing libraries such as FSL for a large cohort in a scalable cloud-based environment. This step is critical when needed to apply and optimize machine learning algorithms at large scale in realistic scenarios. We have compared the performance of three neural network architectures to perform classification of deeper RSNs representing over 50 functional regions, with the MLP providing fastest inference times with a testing accuracy of 100%. This study provides a foundation for an end-to-end rs-fMRI processing and classification pipeline which can be extended to more robust multi-label/hierarchical classifications in the future.

## 6      References

1. Glover, G. H. (2011). Overview of functional magnetic resonance imaging. *Neurosurgery Clinics*, *22*(2), 133-139.
2. Lin, A. L., & Way, H. M. (2014). Functional Magnetic Resonance Imaging. *Pathobiology of Human Disease: A Dynamic Encyclopedia of Disease Mechanisms*, 4005-4018.
3. Nandakumar, N., Manzoor, K., Agarwal, S., Pillai, J. J., Gujar, S. K., Sair, H. I., & Venkataraman, A. (2021). Automated eloquent cortex localization in brain tumor patients using multi-task graph neural networks. *Medical image analysis*, *74*, 102203.

4. Griffanti, L., Douaud, G., Bijsterbosch, J., Evangelisti, S., Alfaro-Almagro, F., Glasser, M. F., ... & Smith, S. M. (2017). Hand classification of fMRI ICA noise components. *Neuroimage*, *154*, 188-205.

5. Kam, T. E., Wen, X., Jin, B., Jiao, Z., Hsu, L. M., Zhou, Z., ... & UNC/UMN Baby Connectome Project Consortium. (2019, October). A deep learning framework for noise component detection from resting-state functional MRI. In *International Conference on Medical Image Computing and Computer-Assisted Intervention* (pp. 754-762). Springer, Cham.

6. Vergun, S., Gaggl, W., Nair, V. A., Suhonen, J. I., Birn, R. M., Ahmed, A. S., ... & Prabhakaran, V. (2016). Classification and extraction of resting state networks using healthy and epilepsy fMRI data. *Frontiers in neuroscience*, *10*, 440.

7. Zhao, Y., Dong, Q., Zhang, S., Zhang, W., Chen, H., Jiang, X., ... & Liu, T. (2017). Automatic recognition of fMRI-derived functional networks using 3-D convolutional neural networks. *IEEE Transactions on Biomedical Engineering*, *65*(9), 1975-1984.

8. Duc, N. T., Ryu, S., Qureshi, M. N. I., Choi, M., Lee, K. H., & Lee, B. (2020). 3D-deep learning based automatic diagnosis of Alzheimer's disease with joint MMSE prediction using resting-state fMRI. *Neuroinformatics*, *18*(1), 71-86.

9. Qureshi, M. N. I., Oh, J., & Lee, B. (2019). 3D-CNN based discrimination of schizophrenia using resting-state fMRI. *Artificial intelligence in medicine*, *98*, 10-17.

10. Nozais, V., Boutinaud, P., Verrecchia, V., Gueye, M. F., Hervé, P. Y., Tzourio, C., ... & Joliot, M. (2021). Deep Learning-based Classification of Resting-state fMRI Independent-component Analysis. *Neuroinformatics*, *19*(4), 619-637.

11. lgayhardt, "What are machine learning pipelines? - Azure Machine Learning." https://docs.microsoft.com/en-us/azure/machine-learning/concept-ml-pipelines (accessed Jul. 09, 2022).

12. M.W. Woolrich, S. Jbabdi, B. Patenaude, M. Chappell, S. Makni, T. Beh-rens, C. Beckmann, M. Jenkinson, S.M. Smith. Bayesian analysis of neuroim-aging data in FSL. NeuroImage, 45:S173-86, 2009

13. Woolrich, M. W., Behrens, T. E. J., Beckmann, C. F., Jenkinson, M., & Smith, S. M. (2004). Multilevel linear modelling for FMRI group analysis using Bayesian inference. NeuroImage, 21(4), 1732–1747

14. L. Nickerson, S.M. Smith, D. Öngür, C.F. Beckmann. Using Dual Regression to Investigate Network Shape and Amplitude in Functional Connectivity Analyses. Front Neurosci. 2017; 11: 115. doi: 10.3389/fnins.2017.00115

15. tamram, "Introduction to Blob (object) storage - Azure Storage." https://docs.microsoft.com/enus/azure/storage/blobs/storage-blobs-introduction (accessed Jul. 09, 2022)

16. "Miniconda — Conda documentation." https://docs.conda.io/en/latest/miniconda.html (accessed Jul. 09, 2022).

17. http://fcon_1000.projects.nitrc.org/fcpClassic/FcpTable.html